\pdfoutput=1

\documentclass[11pt]{article}

\usepackage[final]{acl}

\usepackage{times}
\usepackage{latexsym}

\usepackage[T1]{fontenc}

\usepackage[utf8]{inputenc}

\usepackage{microtype}

\usepackage{inconsolata}

\usepackage{graphicx}
\usepackage{float}
\usepackage{enumitem}

\usepackage{amsmath} 
\usepackage{bbm}

\usepackage[capitalize,noabbrev, nameinlink]{cleveref}

\title{Is That Your Final Answer? Test-Time Scaling \\ Improves Selective Question Answering}

\author{
  William Jurayj \\
 \\
 \\
 \\ \And
  Jeffrey Cheng \\
  Johns Hopkins University \\
  \texttt{\{wjurayj1,jcheng71,vandurme\}@jhu.edu} \\\And
  Benjamin Van Durme \\
  \\
  \\
 \\
}

\makeatletter
\newcommand{\mysubsection}{\subsection}
\newcommand{\myparagraph}{\paragraph}

\makeatother

\begin{document}
\maketitle

\begin{abstract}

Scaling the test-time compute of large language models has demonstrated impressive performance on reasoning benchmarks. However, existing evaluations of test-time scaling make the strong assumption that a reasoning system should always give an answer to any question provided. This overlooks concerns about whether a model is \emph{confident} in its answer, and whether it is appropriate to always provide a response.
To address these concerns, we extract confidence scores during reasoning for thresholding model responses. We find that increasing compute budget at inference time not only helps models answer more questions correctly, but also increases confidence in correct responses. 
We then extend the current paradigm of \textit{zero-risk} responses during evaluation by considering settings with non-zero levels of response risk, and suggest a recipe for reporting evaluations under these settings.\footnote{Code released at \href{https://github.com/wjurayj/final_answer}{https://github.com/wjurayj/final\_answer}}

\end{abstract}

\begin{figure}[t]
    \centering
    \includegraphics[clip, trim=7.2cm 2.4cm 6.8cm 1cm, width=1.0\linewidth]{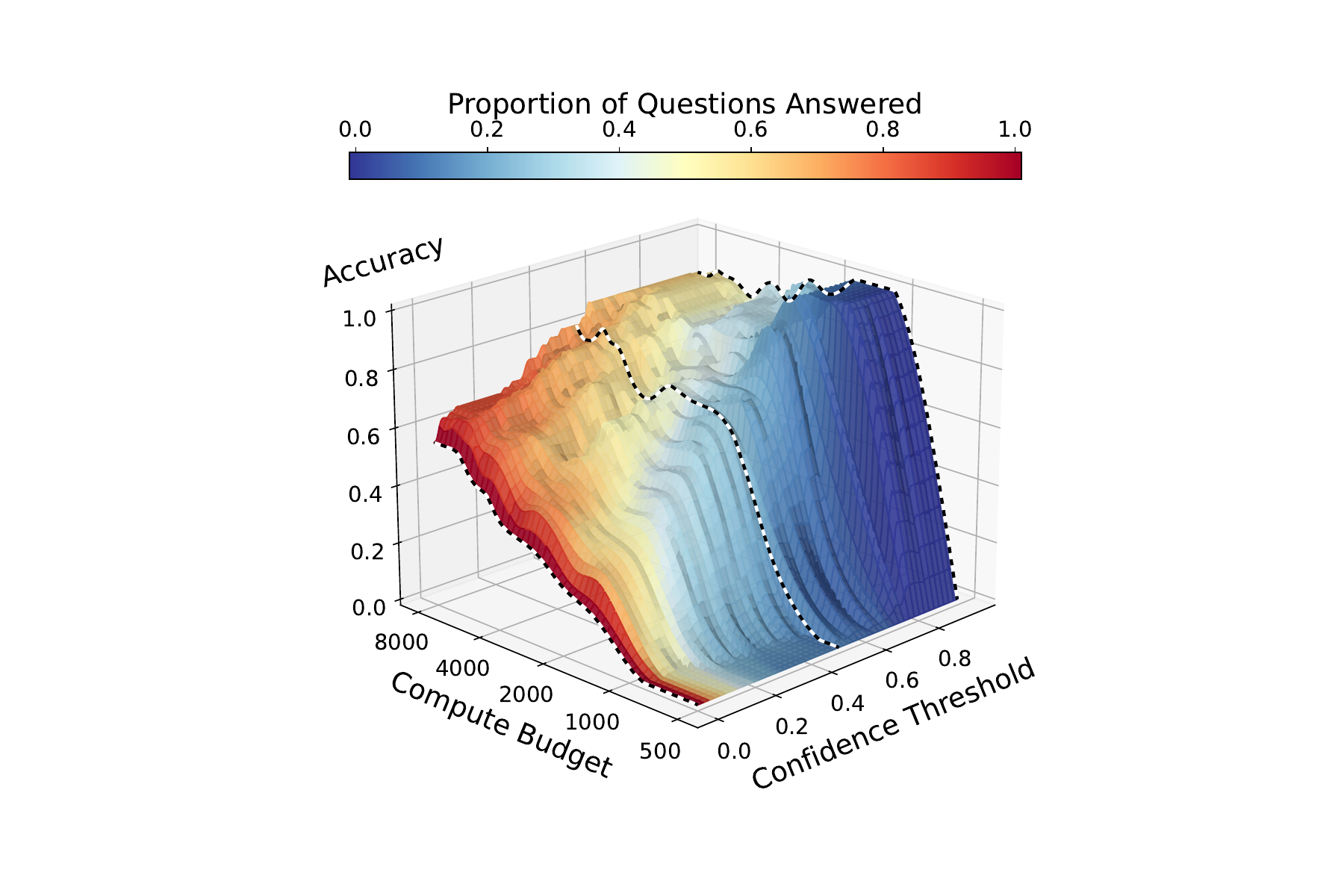}

    \caption{\textbf{DeepSeek R1-32B's accuracy is a function of compute budget and confidence threshold.} Increased confidence thresholds generally yield increased accuracy at the cost of response rate, while increased compute budgets sometimes decrease accuracy as they increase response rate. The vertical axis measures the accuracy of answered questions at a compute budget and confidence threshold. Color indicates the proportion of questions that are answered; in redder regions, the model is more likely to answer, whereas in bluer regions the model is less likely to answer.
    }
    \label{fig:4d-surface}
\end{figure}

\begin{figure*}[t]
    \centering
    \includegraphics[width=\textwidth]{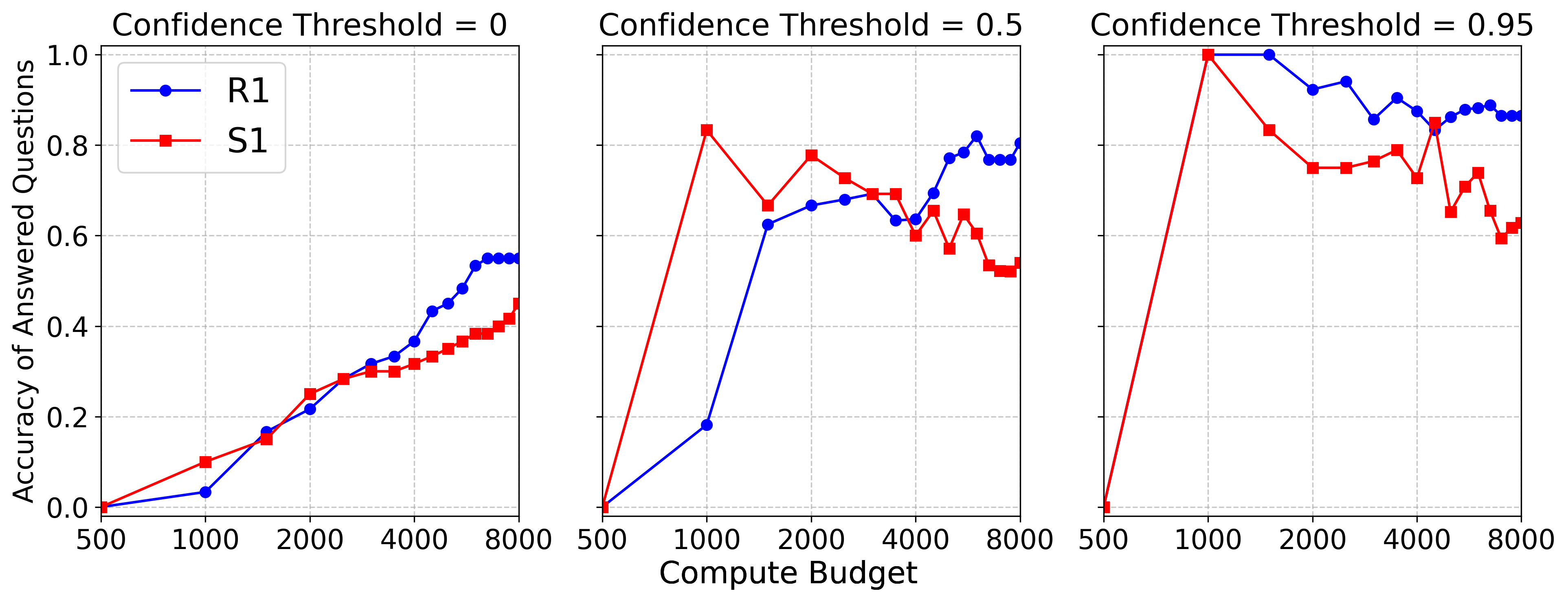}
    \caption{\textbf{Confidence thresholds on test-time scaling.} \textit{(left)} When the confidence threshold is 0, the model answers 100\% of questions. This is the only performance curve that is reported by test-time scaling research. \textit{(center)} At a moderate threshold, more frequent absentions allow higher response accuracy. \textit{(right)} At a high threshold, small amounts of test-time compute deliver very high accuracy at low answer rates, while test-time scaling provides more answers at the cost of answer accuracy.
    }
    \label{fig:threshold}

\end{figure*}

\section{Introduction}
Scaling up language model inference-time compute using lengthy chains of thought has delivered impressive results on mathematical reasoning benchmarks that resisted training compute scaling \citep{deepseek-ai_deepseek-r1_2025, muennighoff_s1_2025}. These results, however, are reported in the zero-risk response setting: with no penalties for incorrect answers, the system always guesses even when it is not confident in its answer. In practice, this behavior is not always desirable.

Many question answering settings associate incorrect answers with measurable costs, ranging from low-risk responses found in game shows \citep{ferrucci_building_2010} to high-stakes responses that can alter people's lives \citep{COMPAS}. \textit{Selective question answering} addresses these challenges by allowing a model to refrain from answering questions which it might answer incorrectly \cite{kamath-etal-2020-selective}.
This requires a selection function, which considers risk tolerance, coverage goals, and candidate answer confidence to decide whether a prediction should be given \cite{geifman_selective_2017}. 
Knowing when not to answer is a critical quality for systems to collaborate effectively with humans \cite{verma_learning_2023}, especially for test-time scaling systems that must constantly decide between refusing to answer and expending further compute to search for a possible solution.

To help address this issue, we evaluate test-time scaling models using a simple class of selection functions that reject questions if a model is not confident in its answer after expending its compute budget. We evaluate these systems at different compute budgets, showing a new axis of model performance that answer accuracy alone struggles to measure. We suggest a class of utility functions that represent various levels of error risk to empirically measure the performance of these systems in settings where incorrect answers are penalized. Evaluation in these settings shows how compute scaling affects confidence in existing systems. Based on these insights, we propose a standard method for measuring model performance in settings with non-zero response risk.
In summary we:
\begin{itemize}[nosep]
    \item Conduct the first evaluation of LLM test-time compute scaling on selective question answering, finding that increasing inference compute can help models distinguish between their correct and incorrect answers. (\cref{sec:experiments})
    \item Introduce evaluation settings that penalize incorrect answers and allow abstentions to help holistically evaluate models capable of scaling test-time compute. (\cref{sec:utility})
    \item Invite the community to report test-time scaling performance on selective question answering under ``Jeopardy Odds'', which incentivize confidence calibration by penalizing incorrect answers while rewarding correct answers.
\end{itemize}

\section{Methods}
We explore how increasing compute budgets affects a model's performance on question answering tasks at different confidence thresholds. The choice of a budget and threshold is a \emph{test-time} decision. We describe methods to quantify the two factors below:

\myparagraph{Compute Budget} refers to the amount of compute expended by the model at inference time. In all cases, we quantify a model's budget by counting the \textit{number of tokens} in its reasoning trace. We use methods proposed by \citet{muennighoff_s1_2025} to strictly enforce compute budgets. Specifically, we ignore any predicted end-of-thinking delimiters and instead append the token ``Wait'' if a model attempts to end its reasoning trace before reaching the budget, and we force decode the end-of-thinking delimiter once the budget is reached.

\myparagraph{Confidence Threshold} refers to the uncertainty of the model in its decoded answer. We quantify a model's confidence as the \textit{sum of the log-probabilities} corresponding to the answer tokens.\footnote{Every answer in our dataset is a 3-digit number between 000 and 999\label{sum}, so consists of the same number of tokens.}
For a confidence threshold, our selection function \cite{geifman_selective_2017} only accepts answers that the model delivers with confidence greater than its threshold, abstaining otherwise.

\begin{figure*}[t]
    \includegraphics[clip, trim=1.1cm 0cm 1cm 1cm, width=1.0\linewidth]{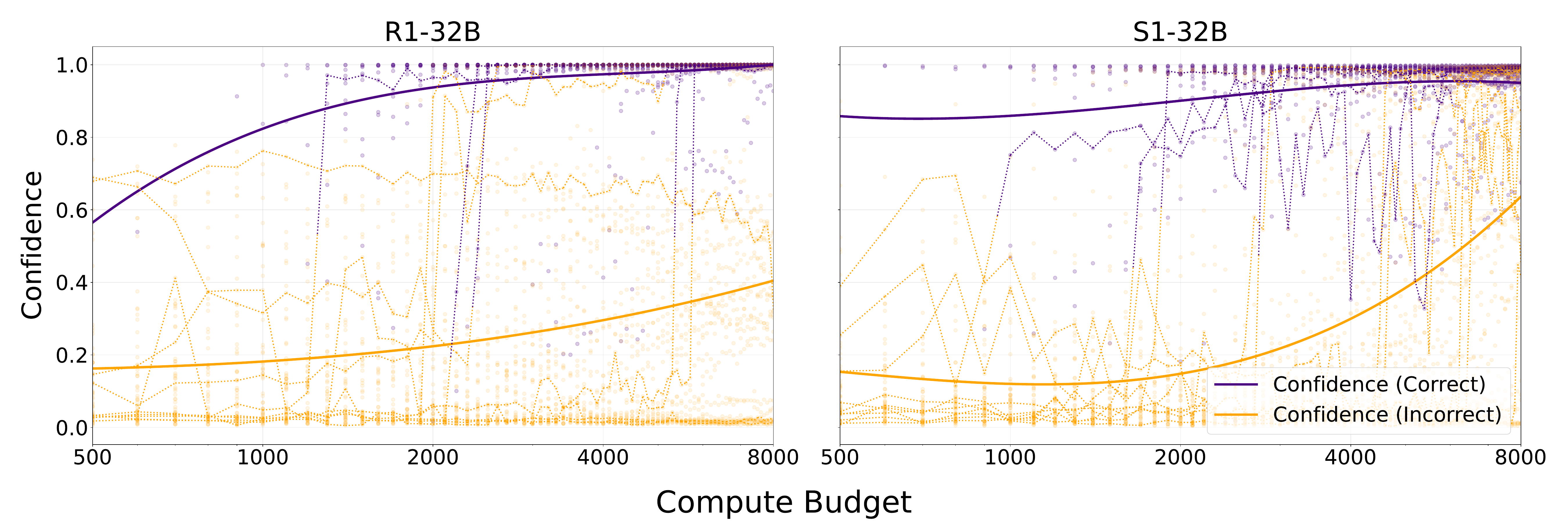}

    \centering
    \caption{\textbf{Test-time scaling improves confidence in correct answers on R1-32B (left) and S1-32B (right).} Each dot represents the model's confidence in an answer after spending a fixed amount of compute. Indigo series are correct answers, while orange series are incorrect. Dotted lines plot confidence trajectories for 10 randomly selected questions, emphasizing how confidence in correct versus incorrect answers changes with test-time compute. Note that individual answers may turn from orange to indigo if the model changes its prediction after thinking longer.}
    \label{fig:confidence}
\end{figure*}

\section{Experiments}
\label{sec:experiments}

\mysubsection{Experimental Setup}

\label{sec:setup}
We evaluate Deepseek-R1-32B \citep{deepseek-ai_deepseek-r1_2025} and s1 \citep{muennighoff_s1_2025} due to their exhibited test-time scaling capabilities and open-weight checkpoints, and choose AIME 2024 and 2025 as our primary evaluation dataset. This dataset contains 60 hard math problems on which performance substantially benefits from larger compute budgets, making it a popular benchmark for evaluating test-time scaling. Additional experiments on GPQA \cite{rein2024gpqa} are included in \cref{sec:GPQA}. We test the set of confidence thresholds $\{0.0, 0.5, 0.95\}$ across compute budgets within the range $[500, 8000]$, incrementing by 100 tokens. For a given budget and threshold, we report the accuracy of \emph{answered} questions, treating never answering as yielding accuracy 0. As the number of answered questions differs across confidence thresholds, we note that accuracies are \emph{not directly comparable} between models and compute budgets.

We use widely available open-source libraries to run our experiments, including HuggingFace Transformers \cite{wolf-etal-2020-transformers} and vLLM \cite{kwon2023efficient} for language model inference, and the Language Model Evaluation Harness
\cite{eval-harness} to sample reasoning chains at temperature 0 and 32-bit precision. In particular, we use the variant of this library released by \citet{muennighoff_s1_2025}, and run a subset of the experiments that they run.
We run experiments on 4 H100 GPUs.

\mysubsection{Results}
\label{sec:results}
\cref{fig:threshold} compares the accuracy of answers provided by R1-32B and S1-32B at different test-time compute budgets. When the confidence threshold is 0, models answer every question, so accuracy increases consistently with compute budget. We observe that these subplots are slices of a surface parameterized by compute budget and confidence threshold, shown in \cref{fig:4d-surface}.
While higher confidence thresholds prevent the model from answering at low budgets, scaling compute at high thresholds delivers a larger volume of accurate answers.
However, at higher confidence thresholds increased compute budget can actually decrease answer accuracy. This decrease in accuracy of yielded answers does not necessarily reflect decreased performance at higher budgets, but instead that the additional questions answered are less likely to be correct than those answered at lower budgets.

To investigate whether excessive thinking harms accuracy drops by pushing models to abandon correct answers, we plot how a model's confidence in individual answers moves over time. \cref{fig:confidence} shows the answer confidences given by both models at varying compute budgets, colored according to their correctness, with a cubic curve fit to the distribution. We note that as compute budget increases, the average confidence of its correct answers increases even as additional correct answers are discovered. Notably, this is not a universal property of test-time scaling models: S1-32B does not separate its correct answers from its incorrect answers as well as R1-32B. 

\section{Utility}
\label{sec:utility}

\mysubsection{Motivation}
\label{sec:motivation}
When refusal to answer is an option, accuracy can be trivially optimized by a system that answers extremely infrequently. Thus, a useful metric must capture both the accuracy of answers provided and the system's propensity to provide answers. Many real world scenarios reward correct answers, but incur measurable costs for incorrect answers. We show our results involving confidence thresholds can be adapted to these settings.

Given a model $\mathcal{M}$ and an instance $x$ of a task $t$, we define a \textit{utility function} $f$ to be 
$$f(\mathcal{M}, x) = \begin{cases}
    1 & \textrm{ $\mathcal{M}$ answers $x$ correctly}\\
    0 & \textrm{ $\mathcal{M}$ abstains from answering $x$ }\\
    r_t  & \textrm{ $\mathcal{M}$ answers $x$ incorrectly}
\end{cases}$$
We can assume the reward for correct answers is 1 without loss of generality due to scaling. While there exist scenarios where refusing to answer also incurs a cost, this paper will only discuss the consequences when no extra cost is incurred; the conclusions we draw can be extended to these cases.

\mysubsection{Problem Scenarios}
\label{sec:scenarios}
We discuss three settings with varying risk levels: 
\begin{itemize}[nosep]
    \item Exam Odds ($r_t = 0$): There are no costs incurred by incorrect answers. These are tasks where guessing isn't punished and the model should always try to provide a solution.
    \item Jeopardy Odds\footnote{Inspired by the wagers made in the game show's `Final Jeopardy' stage} ($r_t = -1$): The cost of an incorrect answer is equal to the reward for a correct answer. In these scenarios, no answer at all is preferable to an incorrect answer.
    \item High-Stakes Odds ($r_t = -20$): The cost of an incorrect answer far outweighs the reward for a correct answer. In this case, the model should answer only if absolutely certain.
\end{itemize}

\begin{figure}[ht]
    \centering
    \includegraphics[clip, trim=7.2cm 2.5cm 6.8cm 4.2cm, width=1.0\linewidth]{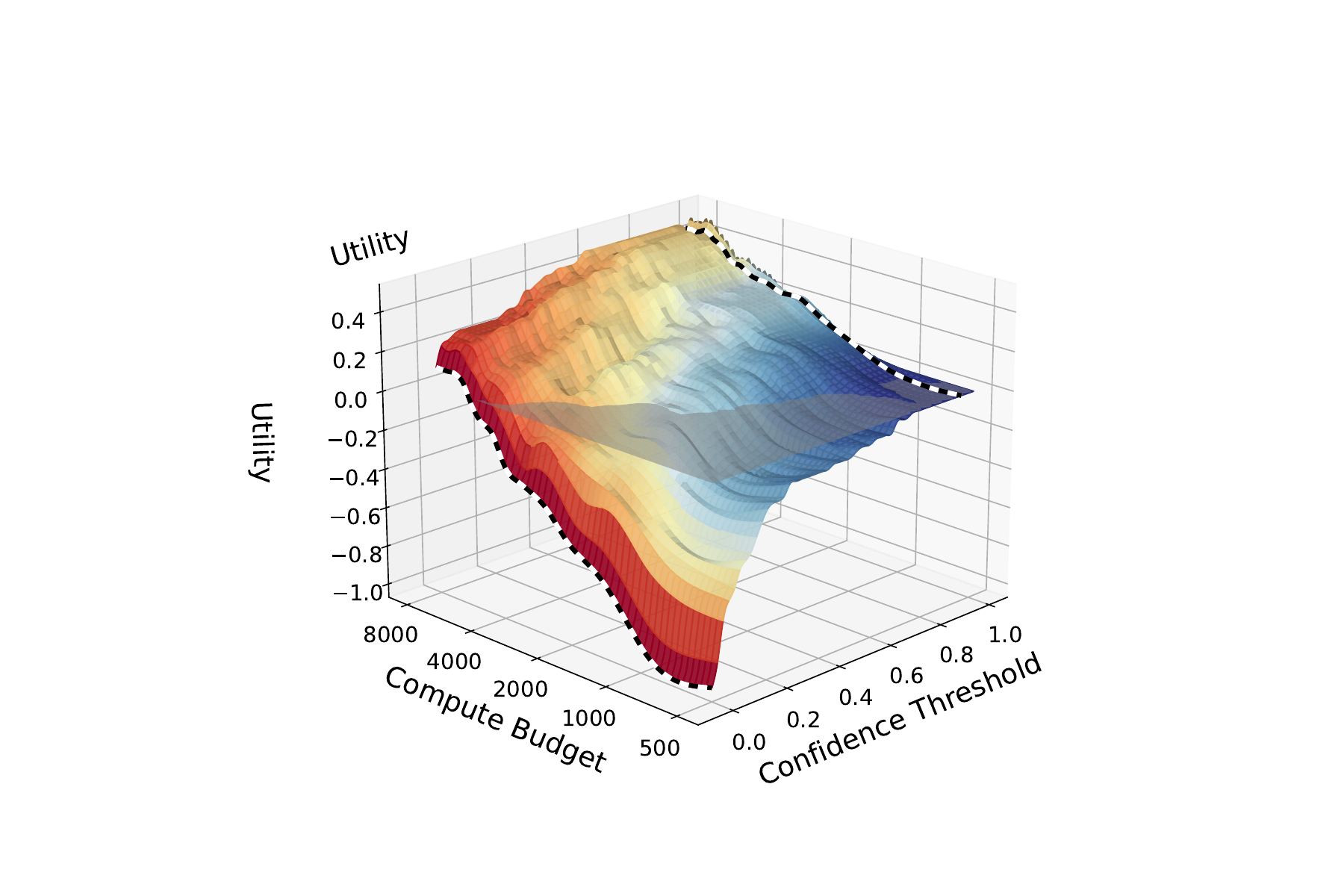}

    \caption{\textbf{Utility Surface of DeepSeek R1-32B for Jeopardy.} The vertical axis indicates performance in the Jeopardy setting at different compute budgets and confidence thresholds. The color indicates the proportion of questions that are answered, as in \cref{fig:4d-surface}. The horizontal plane divides positive and negative utility regions of the operating curve. The checkered lines show the confidence slices that we compare to s1 in \cref{fig:jeopardy}.}
    \label{fig:jeopardy-surface}
\end{figure}

\mysubsection{Results}
\label{sec:results_odds}
We keep the same experimental setup as described in \cref{sec:setup}. Rather than reporting accuracy, we instead report the utility in the three scenarios above, shown in \cref{fig:jeopardy-surface}. We focus on the Jeopardy setting because it highlights why a system might choose not to answer; results in the other settings are in \cref{sec:appendix_figures}.

\begin{figure}[hb]
    \centering
    \includegraphics[width=0.48\textwidth]{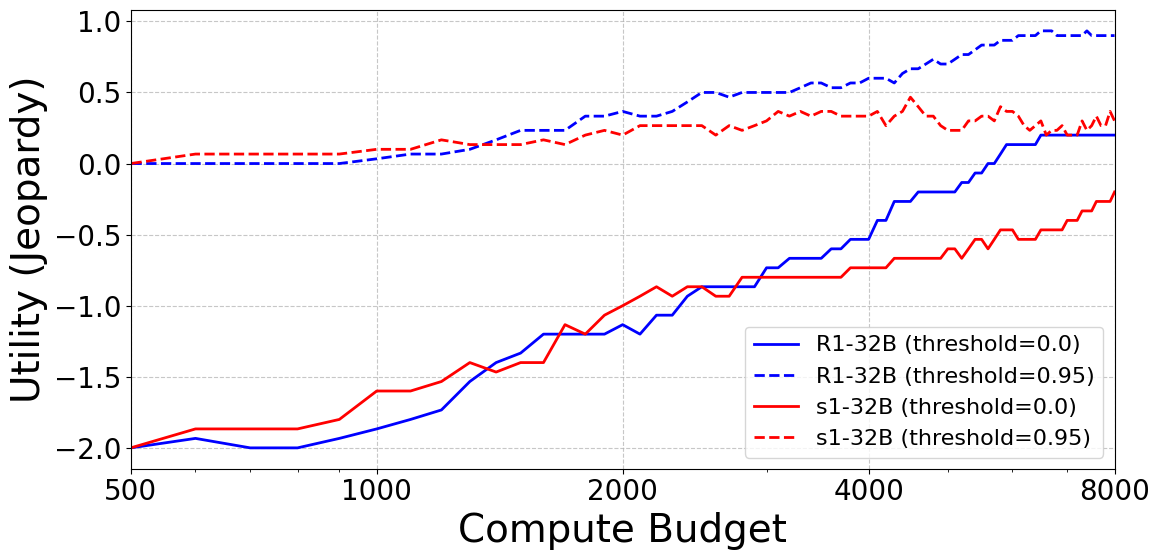}
    \caption{\textbf{Jeopardy utility scales differently across models and thresholds.} Performance of S1-32B and R1-32B in the Jeopardy odds setting under different confidence thresholds. Although S1 is competitive at lower budgets when the confidence threshold is 0, a higher threshold shows R1’s superior scaling performance in a selective setting.}
    \label{fig:jeopardy}
\end{figure}

The Exam setting's utility function does not distinguish refusal from incorrectness, so optimal performance is achieved trivially at confidence threshold to 0 so that every question gets the model's best guess. In the Jeopardy setting, however, this is non-trivial. We illustrate the complete function mapping compute budget and confidence threshold to Jeopardy performance in \cref{fig:jeopardy-surface}: the checkered lines on this surface indicate the two slices that compose R1-32B's portion of \cref{fig:jeopardy}. 
We do not suggest that our choice of 0.95 is the optimal threshold for this task, or even that a threshold is the right approach to confidence calibration. Rather, we apply this naive method to show how test-time scaling for selective classification can benefit a practical question-answering setting.

We see on the left of \cref{fig:threshold} that in the commonly reported Exam Odds, R1-32B and S1-32B scale comparably at threshold 0.
In Jeopardy Odds, selective question answering at threshold 0.95 dramatically improves performance for both models. Additionally, although the two models scale comparably at Exam Odds, R1-32B substantially outperforms S1-32B at larger budgets in this new evaluation setting. Previous work overlooks this comparison.
We call on future test-time compute scaling research to report optimal utility at Jeopardy Odds in addition to Exam Odds, to help readers understand performance across confidence demands.

\section{Related Work}

As scaling training compute has become prohibitively expensive \cite{hoffmann_training_2022}, models that scale performance with test-time compute have become a new frontier \cite{snell2024scalingllmtesttimecompute, wu_inference_2024}. These methods have delivered state-of-the-art results on hard reasoning tasks using lengthy chains of thought \cite{deepseek-ai_deepseek-r1_2025, muennighoff_s1_2025}. Current work in this space optimizes for question answering tasks which do not penalize incorrectness, ignoring settings that favor refusal over wrong answers \cite{ferrucci_building_2010, rajpurkar_know_2018, kamath-etal-2020-selective}.
We draw motivation from methods for cost-sensitive learning \cite{mienye_performance_2021} and selective classification \cite{geifman_selective_2017}, which navigate penalties for failure. These settings reward confidence calibration, which can be critical for effective collaboration with human experts \cite{verma_learning_2023}. We are the first to investigate how serialized test-time compute helps models identify when they should not answer.

\mysubsection{Test-time Scaling}
Many methods for scaling test-time compute have been explored. These include searching over possible generations \citep{wang2024hypothesissearchinductivereasoning}, sampling many completions and selecting the best answer among them \cite{wu_inference_2024}, making gradient updates at inference time \cite{akyurek_surprising_2024, li_combining_2024}, using reinforcement learning to incentivize generating chains of thought before answering \cite{deepseek-ai_deepseek-r1_2025}, and simply fine-tuning on longer chains of thought \citep{muennighoff_s1_2025}.
Our work considers models fine-tuned on very long reasoning chains, and augments them with the ability to refuse to answer questions where they lack confidence. Concurrent work finds that such models can learn to predict when their answers are unlikely to be correct \cite{zhang2025reasoningmodelsknowtheyre, huang2025efficienttesttimescalingselfcalibration}, but do not show how this affects performance as additional compute is expended. In contrast, we show how model confidence scales with test-time compute, and demonstrate its value for optimizing performance in settings that allow refusal to answer.

\mysubsection{Selective Question Answering}
Refusing to answer is an important option in many prior works on question answering. SQuAD 2.0 included this feature by asking questions which have no answer, although they treat abstaining from answering as a correct answer in these unanswerable cases \cite{rajpurkar_know_2018}. Game-show based research efforts use an approach more closely aligned with ours, which penalizes systems for answering incorrectly to encourage abstentions when a system cannot develop sufficiently high confidence \cite{ferrucci_building_2010, ferrucci_introduction_2012, boyd-graber-borschinger-2020-question, rodriguez_quizbowl_2021}. Related to quiz game settings is research into selective classification, which evaluates models performance across the coverage-accuracy curve, rather than at single point \cite{geifman_selective_2017}. These approaches can be useful for avoiding costly errors in high-pressure domains \cite{khan_cost_2017}, under distribution shift \cite{ren_out--distribution_2023}, or when designing systems that defer to expert humans when it lacks confidence that its input will be helpful \cite{mozannar_consistent_2021}.
Recent research in language modeling has investigated training language models to refuse to answer \cite{cao_learn_2024}, and this capacity for refusal has become a point of competition among top industrial labs \cite{wei2024measuringshortformfactualitylarge}.
However, this line of work does not investigate this behavior in sequential test-time scaling models on reasoning intensive tasks, where a model might find a confident answer given higher compute budgets.

\section{Conclusion}

We highlight a region of performance that is currently unexplored by test-time scaling research.
We encourage the test-time scaling community to adopt these insights by reporting model scaling performance on benchmarks at both Exam Odds and Jeopardy Odds, to highlight their systems ability to scale confidence with test-time compute. Future work should focus on efficiently allocating test-time compute to meet confidence demands, and could investigate how test-time confidence scaling models should decide between extending reasoning and deferring to human experts.

\section*{Limitations}

The selection function we implement is based entirely on the likelihood that a large language model assigns a series of tokens after thinking, which is not necessarily the optimal method for model confidence estimation.
Furthermore, the Chain-of-Thought scaling method we apply may struggle to generalize to problem types that a model has note
The method we use for `budget forcing' \citep{muennighoff_s1_2025} may diminish performance by abruptly truncating chains of thought and driving the model outside of its training distribution. Concurrent work has introduced more elegant forms of compute budget control \cite{aggarwal_l1_2025, hou_thinkprune_2025}.
Furthermore, we do not consider how compute costs might be incorporated in the model's utility function, which could encourage increased energy consumption.
Finally, we recognize that by evaluating only on English questions and answers, we may miss model capabilities or weaknesses in lower-resource languages or in multilingual settings.

\section*{Acknowledgments}

This work was funded in part by the U.S. National Science Foundation under grant No. 2204926, and by the Defense Advanced Research Project Agency (DARPA) SciFy program (contract No. HR001125C0304) and CODORD program.
Any opinions, findings, and conclusions or recommendations expressed in this material are those of the author(s) and do not necessarily reflect the views of the National Science Foundation or DARPA.
We thank Miriam Wanner, Zhengping Jiang, Orion Weller, Marc Marone, Alex Martin, and Beepul Bharti for helpful conversations throughout this project.

\bibliography{acl_latex}
\newpage
\appendix

\section{Appendix}
\label{sec:appendix_figures}

\begin{figure}[H]
    \centering
    \includegraphics[clip, trim=7.2cm 2.4cm 6.8cm 1cm, width=1.0\linewidth]{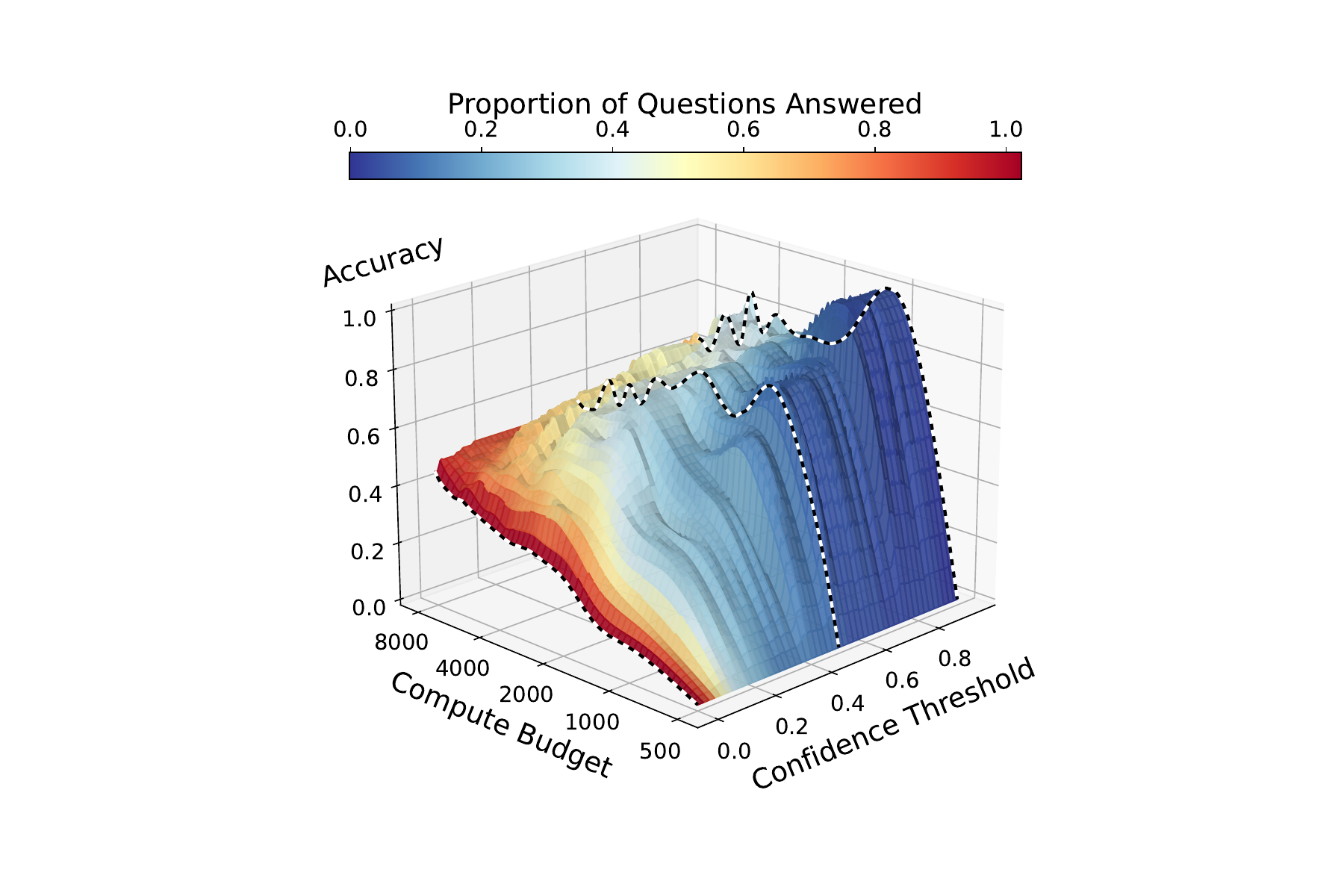}

    \caption{\textbf{(above) S1-32B's answer accuracy is a function of compute budget and confidence threshold.} This plot corresponds to the R1-32B plot in \cref{fig:4d-surface}.
    \textbf{(below) Utility surface of S1-32B for Jeopardy.} This plot corresponds to the R1-32B plot in \cref{fig:jeopardy-surface}.
    }
    \label{fig:4d-surface-s1}
    \includegraphics[clip, trim=7.2cm 2.5cm 6.8cm 4.2cm, width=1.0\linewidth]{figures/jeopardy_surface_r1.pdf}

\end{figure}

\begin{figure}
    \centering
    \includegraphics[width=0.48\textwidth]{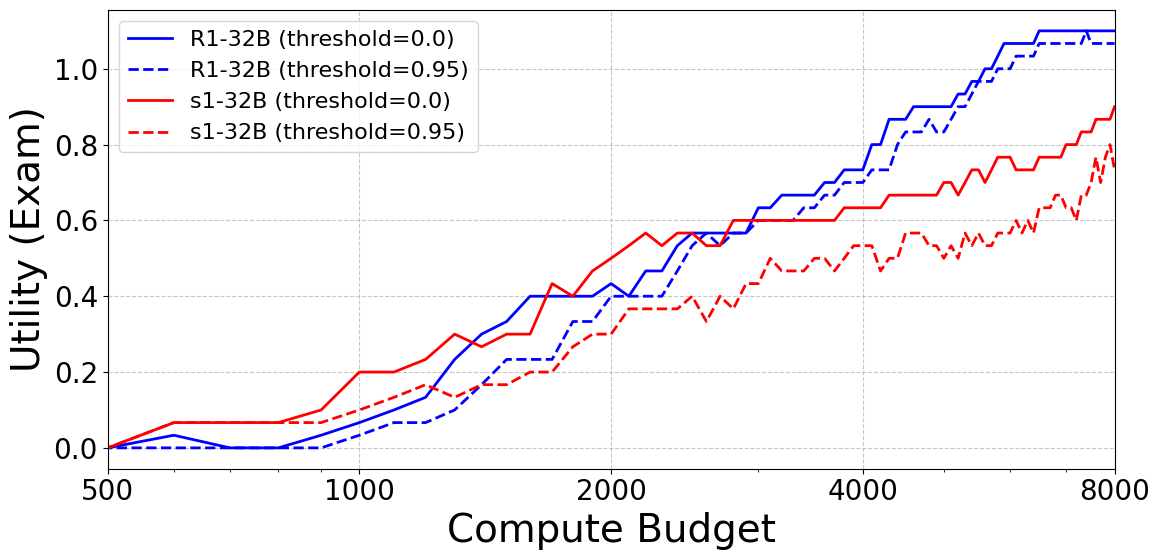}
    \caption{\textbf{Additional Model Comparisons.} We additionally compare performance of S1-32B and R1-32B in the Exam Odds \textit{(above)} and High-Stakes Odds \textit{(below)} settings under different confidence thresholds. Like Jeopardy odds depicted in \cref{fig:jeopardy}, High-Stakes Odds illustrates a performance distinction at high confidence thresholds that is not evident from conventional Exam odds.}
    \includegraphics[width=0.48\textwidth]{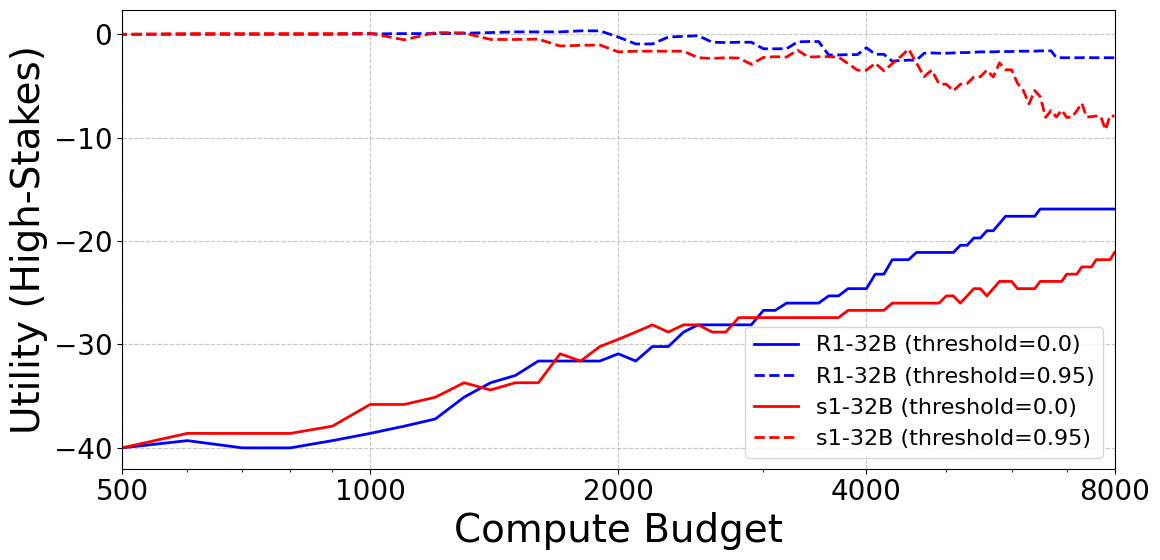}
    \label{fig:odds}
\end{figure}

\section{GPQA Experiments}
\label{sec:GPQA}

GPQA consists of 448 graduate level multiple choice questions in the fields of physics, biology, and chemistry. These questions are considered ``Google-proof'', in that skilled non-experts with access to the open internet struggle to answer them. We run experiments on the `Diamond' subset of this dataset, which consists of the 198 questions which had the clearest answers to domain experts, while being the most difficult for non-experts to answer; this subset has also served as a benchmark for test-time scaling language models \cite{muennighoff_s1_2025}. Our experiments on GPQA follow the same basic procedure described in \cref{sec:experiments}, except that we stop evaluate token budgets in range [500, 4000].

\begin{figure*}[t]
    \centering
    \includegraphics[width=\textwidth]{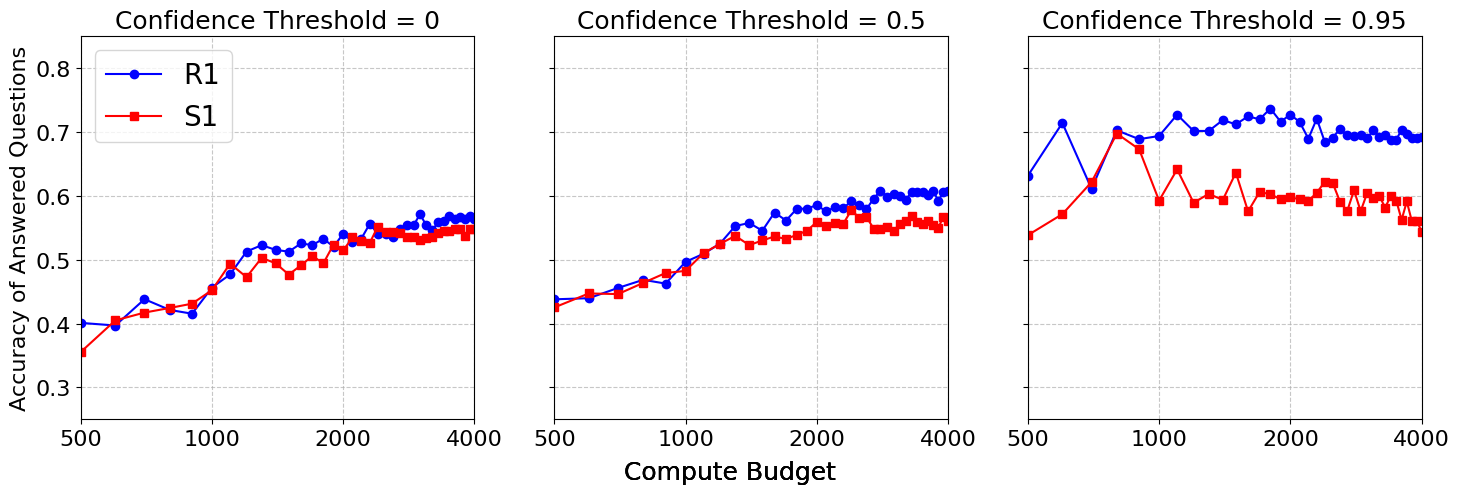}
    \caption{\textbf{Confidence thresholds on test-time scaling.} \textit{(left)} When the confidence threshold is 0, the model answers 100\% of questions. This is the only performance curve that is reported by test-time scaling research. \textit{(center)} At a moderate threshold, more frequent absentions allow higher response accuracy. \textit{(right)} At a high threshold, small amounts of test-time compute deliver very high accuracy, while test-time scaling provides more answers at the cost of answer accuracy. We treat the decision to never answer as yielding accuracy 0.}
    \label{fig:threshold_gpqa}
\end{figure*}

As with AIME, \cref{fig:threshold_gpqa} shows how models can increase performance by only answering when highly confident. Notably, there is very little gap between thresholds 0 and 0.5 on GPQA, owing to its multiple choice format. Moreover, whereas R1-32B outperformed S1-32B at all confidence thresholds, R1-32B only exceeds S1-32B at a confidence threshold of 0.95. 

\begin{figure}[H]
    \centering
\    \includegraphics[clip, trim=7.2cm 2.4cm 6.8cm 1cm, width=1.0\linewidth]{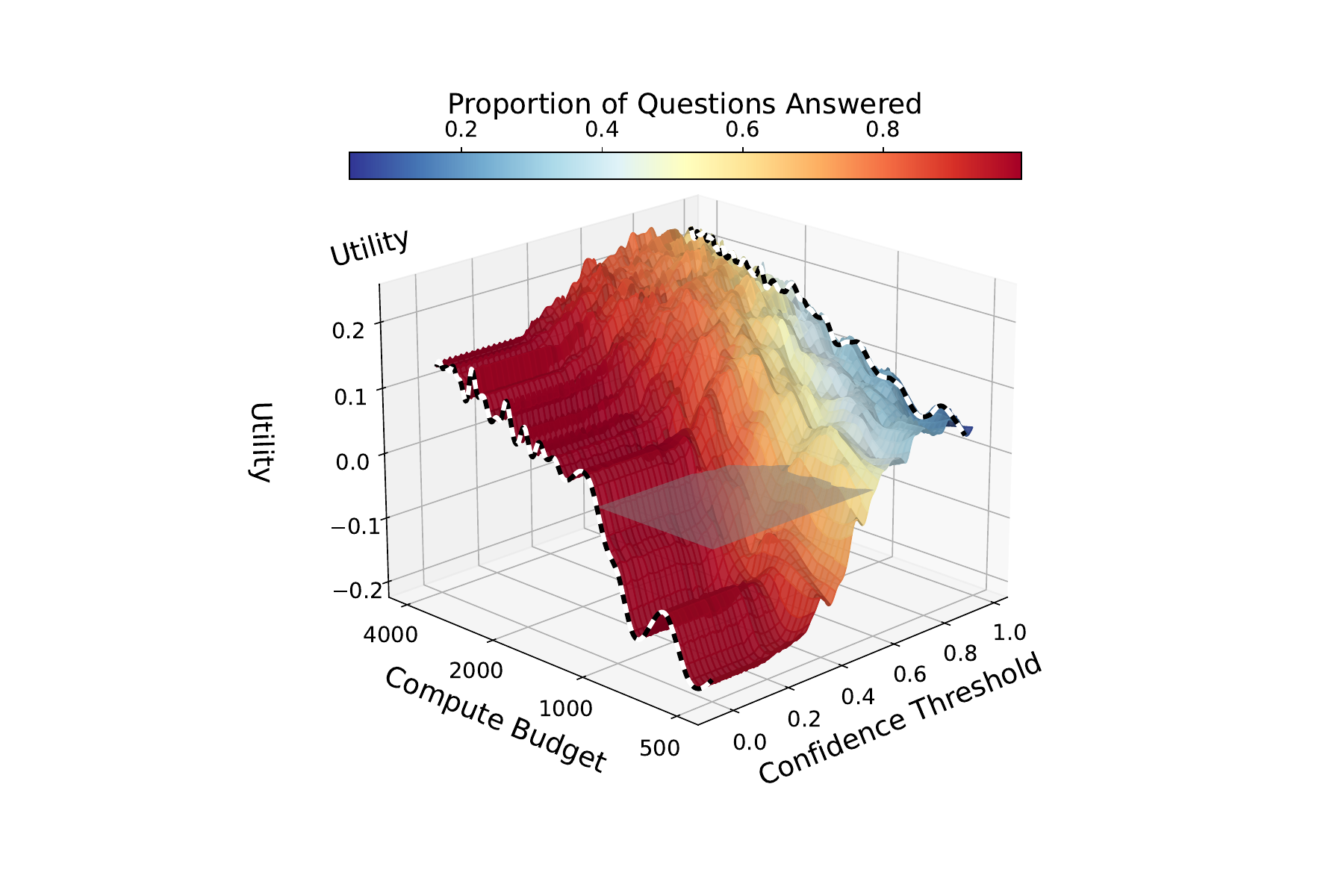}

    \caption{\textbf{Utility surfaces of R1-32B (above) and S1-32B (below) for Jeopardy utility on GPQA}. Utility is a function of compute budget and confidence threshold. These plots mirror the surfaces in \cref{fig:jeopardy-surface} and \cref{fig:4d-surface-s1}.}
    \includegraphics[clip, trim=7.2cm 2.5cm 6.8cm 4.2cm, width=1.0\linewidth]{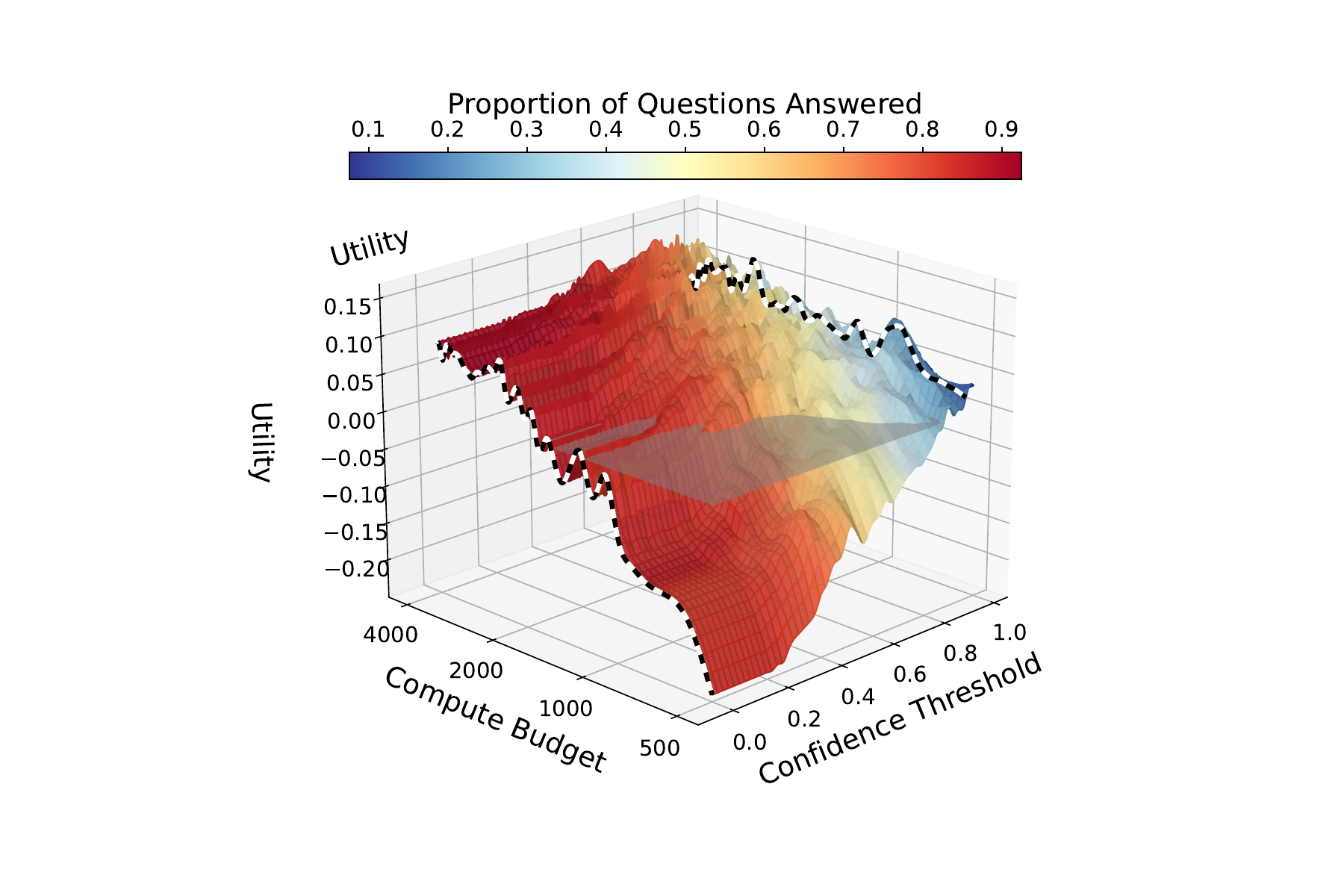}
    \label{fig:jeopardy-surface-gpqa}

\end{figure}

\begin{figure}[hb]
    \centering
    \includegraphics[width=0.48\textwidth]{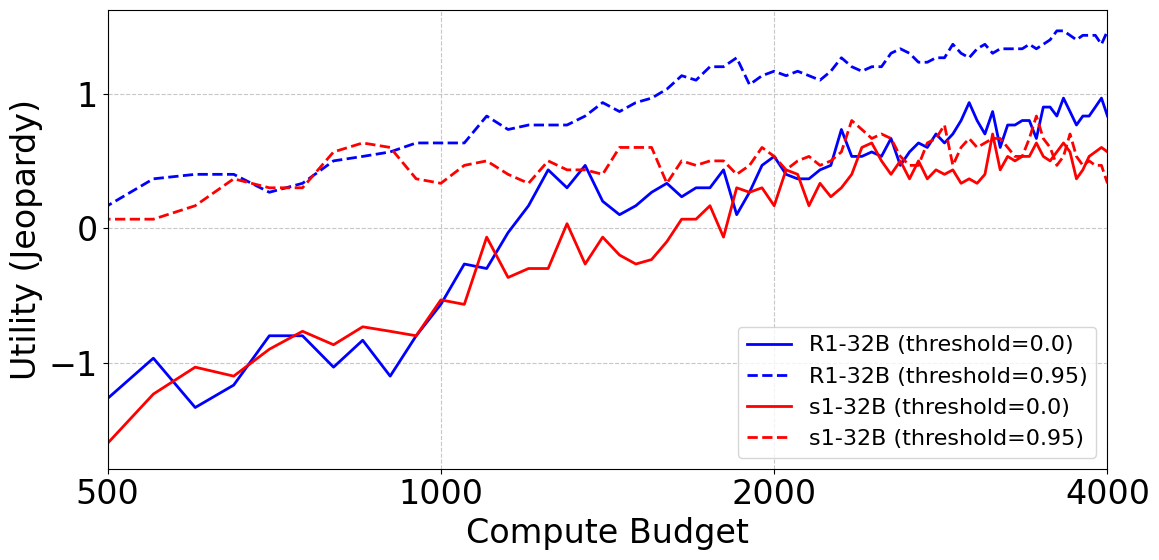}
    \caption{\textbf{Jeopardy utility on GPQA scales differently across models and thresholds.} Performance of S1-32B and R1-32B on GPQA in the Jeopardy odds setting under different confidence thresholds. While S1 is competitive in the case when threshold is 0, a higher threshold shows R1’s superior scaling performance.}
    \label{fig:jeopardy_gpqa}
\end{figure}

\cref{fig:jeopardy-surface-gpqa} shows performance at Jeopardy odds of R1-32B (above) and S1-32B (below). Although both models perform similarly at low thresholds, R1-32B's superior test-time scaling behavior becomes evident at higher confidence thresholds, allowing it to achieve higher peak utilities compared to S1-32B. This gap is reflected in the slices shown in \cref{fig:jeopardy_gpqa}.
We also note that the slices of the surface are nearly identical at lower confidence thresholds below 0.25. This is likely due to the multiple choice format of GPQA; when the model does not know the answer to a question, it assigns roughly equal probability to the four possible answers.

\end{document}